\documentclass[letterpaper, 10 pt, conference]{ieeeconf}  

\IEEEoverridecommandlockouts                              

\usepackage[utf8]{inputenc}
\usepackage[T1]{fontenc}
\usepackage{mathptmx}
\usepackage{times}
\usepackage{graphics}
\usepackage{graphicx}
\usepackage{epsfig}
\usepackage{amsmath, amssymb}
\usepackage{hyperref}
\usepackage{booktabs}
\usepackage{xcolor}
\usepackage{multirow}
\usepackage{array}
\let\labelindent\relax
\usepackage{enumitem}
\usepackage{url}
\usepackage{siunitx}
\usepackage{pifont}
\usepackage{caption}
\usepackage{subcaption}
\usepackage{microtype}
\usepackage{cite}
\usepackage{minted}
\usepackage{cuted}   
\usepackage{capt-of} 
\usepackage{flushend}
\usepackage{xspace}

\newcommand{\projectpage}{\href{https://m80hz.github.io/kite/}{\texttt{https://m80hz.github.io/kite/}}\xspace}

\title{KITE: Keyframe-Indexed Tokenized Evidence for VLM-Based Robot Failure Analysis}
\author{Mehdi Hosseinzadeh, King Hang Wong, and Feras Dayoub
\\
\projectpage
\thanks{Authors are with the Australian Institute for Machine Learning (AIML), Adelaide University, Australia.}
}

\begin{document}
\maketitle

\thispagestyle{empty}
\pagestyle{empty}


\begin{strip}
  \vspace*{-1.5\baselineskip} 
  \centering
  \includegraphics[width=\textwidth]{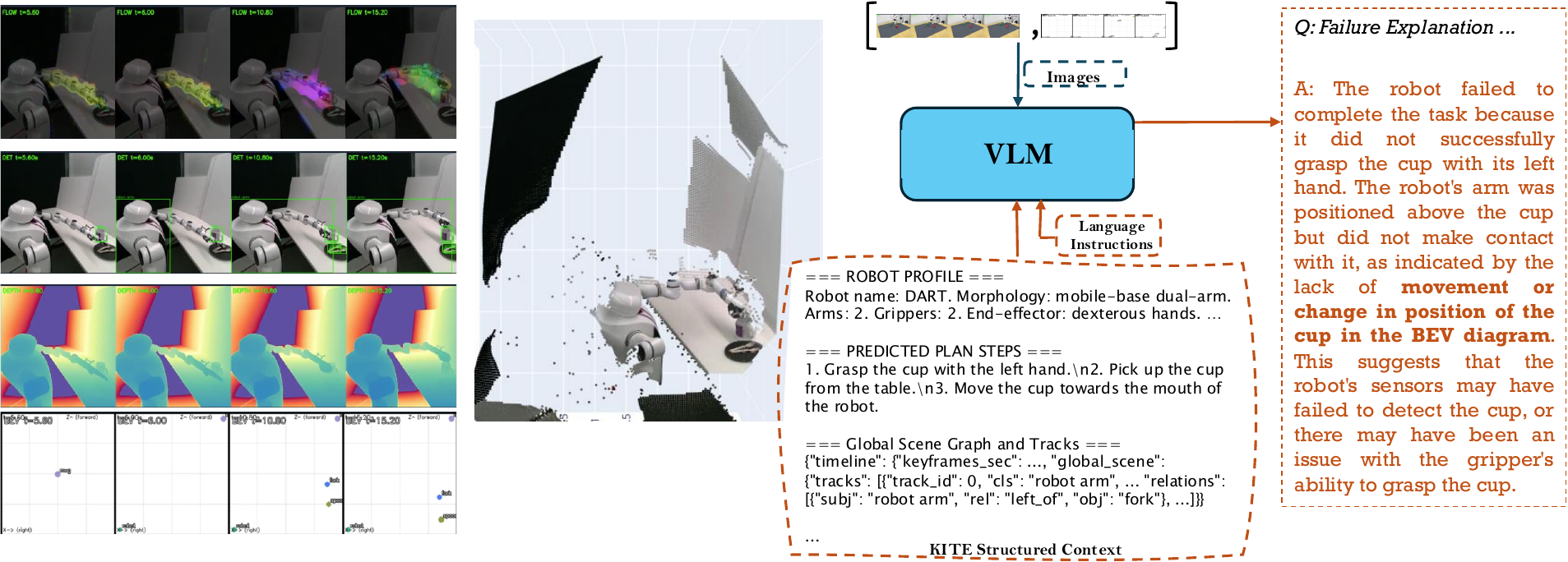}
  \captionof{figure}{\textbf{Failure explanation in real-world settings with KITE.}
  Example sequence from the Dual-Arm Robot (DART) in the lab. Left (top to bottom):
  optical flow estimates used for keyframe selection; RGB keyframe with object detection overlays;
  single-view depth estimates; and a pseudo-BEV schematic (circles with radius $\propto$ confidence;
  X/Z axes; timestamp). Notably, in its failure explanation, the VLM references the BEV diagram to
  infer that the cup's position remains unchanged.}
  \label{fig:teaser}
  \vspace*{-0.5\baselineskip} 
\end{strip}


\begin{abstract}
We present KITE, a training-free, keyframe-anchored, layout-grounded front-end that converts long robot-execution videos into compact, interpretable tokenized evidence for vision-language models (VLMs). KITE distills each trajectory into a small set of motion-salient keyframes with open-vocabulary detections and pairs each keyframe with a schematic bird's-eye-view (BEV) representation that encodes relative object layout, axes, timestamps, and detection confidence. These visual cues are serialized with robot-profile and scene-context tokens into a unified prompt, allowing the same front-end to support failure detection, identification, localization, explanation, and correction with an off-the-shelf VLM. On the RoboFAC benchmark, KITE with Qwen2.5-VL substantially improves over vanilla Qwen2.5-VL in the training-free setting, with especially large gains on simulation failure detection, identification, and localization, while remaining competitive with a RoboFAC-tuned baseline. A small QLoRA fine-tune further improves explanation and correction quality. We also report qualitative results on real dual-arm robots, demonstrating the practical applicability of KITE as a structured and interpretable front-end for robot failure analysis. Code and models are released on our project page: \url{https://m80hz.github.io/kite/}
\end{abstract}


\section{Introduction}
Robots executing long-horizon manipulation in the wild still fail in mundane but consequential ways: a gripper approaches a mug off-axis and slides off, a handle is contacted too late relative to the arm motion, or a bimanual handover misaligns in space and time. Explaining such failures requires combining \emph{where} (layout, contact, relative pose), \emph{when} (the moment the execution deviates), and \emph{what} (task intent and subgoals). In practice, these cues are often subtle, distributed across time, and difficult to recover from raw video alone.

This challenge helps explain why the robotics community has increasingly turned to large language models (LLMs) and vision-language models (VLMs) as general-purpose reasoning interfaces. Foundation models offer open-vocabulary perception, natural-language conditioning, reusable commonsense priors, and a single interface that can support planning, monitoring, explanation, and recovery across many tasks and embodiments \cite{brohan2023can,driess2023palm,reed2022generalist,hu2023toward,firoozi2023foundation,liu2024moka,huang2024copa,huang2024rekep}. Yet their strengths are blunted when the input is a long raw execution video: subtle failure cues are easily buried in dense visual detail, temporal context is diluted, and the evidence needed for diagnosis is rarely presented in a form that is immediately legible to the model.

Prior work has begun to address this problem by summarizing robot experiences for an LLM \cite{liu2023reflect}, training failure-specific VLMs \cite{duan2024aha}, or introducing failure-analysis QA benchmarks \cite{robofac2025}. These directions are important, but they either depend on additional task-specific training, or they still leave open the representation problem: how should a long execution be converted into a compact form that preserves the key spatiotemporal evidence needed for post-hoc diagnosis? To the best of our knowledge, there is still no simple training-free front-end that makes those facts immediately accessible to an off-the-shelf generalist VLM.

To address this gap, we propose \emph{Keyframe-Indexed Tokenized Evidence} (KITE), a compact and interpretable front-end that converts an execution into a small set of motion-salient keyframes, each paired with a schematic pseudo-BEV of relative layout and a minimal scene/interaction summary. As previewed in Fig.~\ref{fig:teaser}, KITE externalizes the evidence a VLM needs: object detections, coarse depth ordering, robot-profile information, scene relations, timestamps, and contact-transition cues are turned into a temporally indexed storyboard and serialized prompt context. The resulting representation supports failure detection, identification, localization, explanation, and correction without task-specific prompt redesign.

In summary, the contributions of this paper are as follows: 
\begin{itemize}
\item A training-free, keyframe-indexed, layout-grounded front-end that converts long robot-execution videos into compact evidence consumable by general-purpose VLMs;
\item A multimodal evidence representation that combines pseudo-BEV schematics, robot-profile information, scene relations, and contact-transition cues in a single interpretable prompt format;
\item A keyframe-indexed failure localization method, in contrast to approaches that localize failures only at coarse plan steps.
\end{itemize}
On RoboFAC~\cite{robofac2025}, KITE with a general-purpose VLM substantially improves over vanilla Qwen2.5-VL and remains competitive with a RoboFAC-tuned baseline, while a lightweight QLoRA fine-tune further improves explanation and correction quality. We also demonstrate qualitative effectiveness on rollout episodes from two real dual-arm robots, the RealMan Dual Arm Compound Robot \cite{realman2024compound} and ALOHA-2 \cite{aloha2}.

\section{Related Work}
\label{sec:related}

\textbf{Foundation Models for Robotics.} 
Recent advances in Large-Language-Models (LLMs) \cite{zeng2023large,zhang2023large,achiam2023gpt,touvron2023llama} and Vision-Language-Models (VLMs)~\cite{openai2024gpt4o,liu2023llava,liu2024llavanext,reid2024gemini,team2023gemini} have catalyzed their integration into robotics across planning, control, and interaction domains. Numerous works leverage pretrained LLMs as high-level planners for robots, using natural language understanding and commonsense to decompose tasks and guide actions. For example, \cite{brohan2023can} pioneered grounding an LLM in robotic affordances for instruction-following, and subsequent systems \cite{duan2022survey,hu2023toward,firoozi2023foundation,driess2023palm,reed2022generalist,ouyang2022training,crosby2013automated,xu2023rewoo,kojima2022large} have combined language and perception in embodied models that reason over visual inputs to plan robot behavior \cite{liu2024moka, huang2024copa, huang2024rekep}. LLM-driven policies have been applied to navigation and mobile manipulation tasks, including domestic assistive robots that follow instructional/guiding prompts to tidy environments or perform user requests. Such LLM-based planners and embodied agents have demonstrated flexible task generalization and improved semantic understanding in novel scenarios. 
This broader shift motivates our setting: if foundation models are increasingly used to plan and guide robot behavior, they should also be able to analyze and explain robot failures. The key difficulty is that post-hoc failure analysis requires evidence to be presented in a temporally compact and spatially legible form.

\textbf{Robotic Failure Analysis, Retrospection, and Recovery.}
Failure explanation has long been studied in explainable robotics and human-robot interaction (HRI), including verbalization, user-facing explanations, and recovery support \cite{das2021explainable,rosenthal2016verbalization,ye2019human,khanna2023user,arkin2020multimodal,bucker2022latte}. More recently, LLM-based methods have been used to summarize experiences, diagnose failures, and suggest corrective actions \cite{liu2023reflect,raman2024cape,wang2024can,dechant2023learning}. In parallel, VLM-based approaches have been explored as success or failure detectors, and instruction-tuned models have been trained specifically for robotic failure reasoning \cite{duan2024aha,du2023vision,ma2022vip,ha2023scaling,wang2023gensim}. These methods show the promise of foundation-model reasoning, but many rely on task-specific fine-tuning or large memory structures.

\textbf{Structured Representations for Multimodal Reasoning.}
Recent work suggests that structured scene abstractions can inject useful inductive bias into language-guided reasoning without requiring full retraining. Examples include 3D scene graphs for task grounding \cite{rana2023sayplan}, bird's-eye-view (BEV) interfaces for multimodal reasoning \cite{choudhary2024talk2bev}, and diagrammatic abstractions for improved visual understanding \cite{deng2021sketch}. KITE is most closely aligned with this direction. Compared with REFLECT \cite{liu2023reflect}, which reasons over summarized robot memories and multisensory logs, KITE focuses on the representation interface itself: it converts long execution videos into a compact, keyframe-indexed, layout-grounded evidence bundle that can be consumed directly by a pretrained VLM. Compared with tuning-heavy failure-analysis approaches \cite{duan2024aha}, KITE remains training-free at the front end and is designed to preserve interpretability through explicit keyframes, schematic layouts, and serialized evidence tokens.

\section{Method: KITE Front-End}
\label{sec:method}

\begin{figure*}[t!]
\centering
\includegraphics[width=\linewidth]{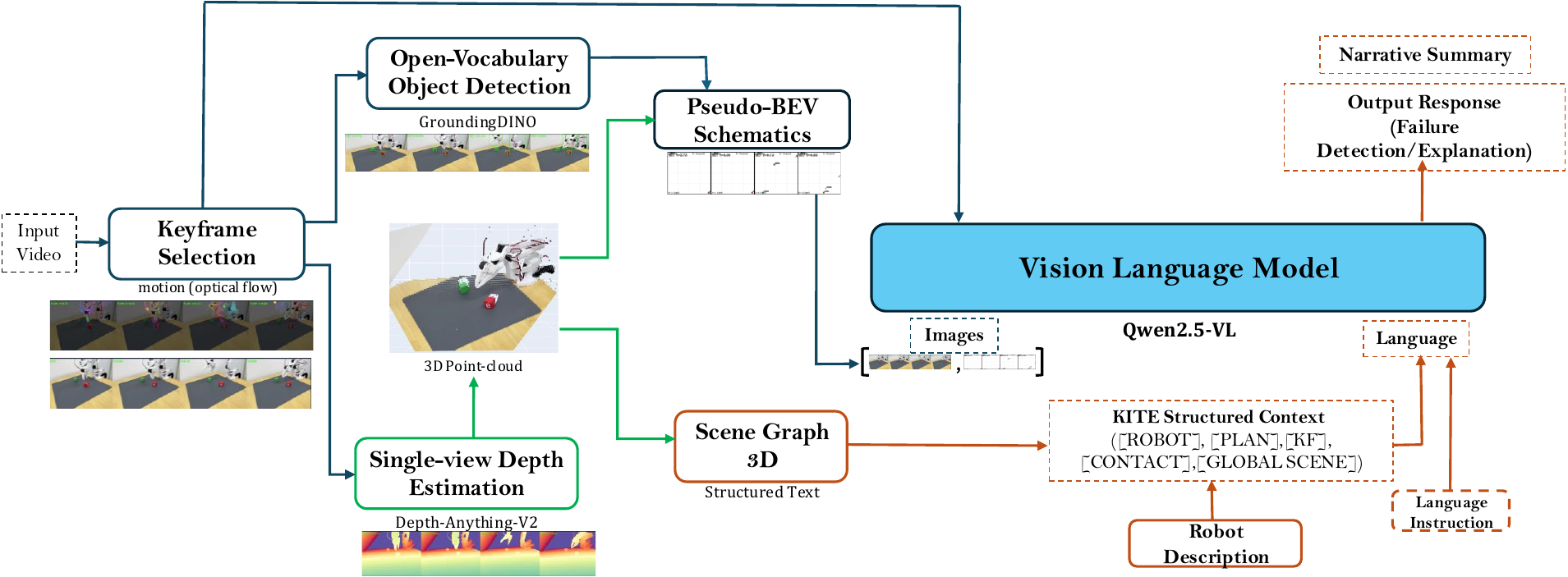}
\caption{\textbf{Overview of KITE.} The proposed pipeline takes a raw video and distills it into a small set of salient keyframes, identified using motion-based peaks. For each keyframe, we run open-vocabulary detection to localize the robot and surrounding objects, and render a pseudo-BEV schematic that depicts the scene layout with simple, interpretable symbols. These visual elements are paired with a structured context and form a compact, interpretable front-end for prompting a vision-language model. The model can then answer diverse failure analysis QA tasks, as well as generate grounded explanations and final narratives. The sequence illustrated here comes from the real-world subset of the RoboFAC benchmark~\cite{robofac2025}.}
\label{fig:overview}
\end{figure*}

\noindent
\textbf{KITE} is a \emph{training-free}, model-agnostic front-end that converts a long robot-execution video into a compact bundle of motion-salient keyframes, schematic pseudo-BEVs, and serialized evidence tokens. Each component is designed to externalize a factor that is otherwise difficult for a VLM to infer reliably from raw video alone: \emph{when} via keyframes, \emph{where} via pseudo-BEV and scene relations, and \emph{what/how} via robot-profile and task-context tokens. The resulting representation is layout-grounded, temporally indexed, and directly consumable by a general-purpose VLM. An overview appears in Fig.~\ref{fig:overview}.

\subsection{Preliminaries and Notation}
Let a video be $\mathcal{V}=\{(I_t,t)\}_{t=1}^{T}$, where $I_t \in \mathbb{R}^{H\times W\times 3}$ is an RGB frame and $t$ is its timestamp. We select up to $M$ keyframes
\[
K=\{(I_{t_k}, t_k, i_k)\}_{k=1}^{M},
\]
where $i_k$ is the frame index. For each keyframe $k$, we compute: (i) an open-vocabulary detection set $O_k=\{(b_j,c_j,s_j)\}$ with box $b_j$, class $c_j$, and confidence $s_j$; and (ii) a relative depth map $D_k$ (single-view, up to scale). For each consecutive keyframe pair $(k,k+1)$, we additionally compute a contact-transition token
\[
\gamma_{k\rightarrow k+1} \in \{\textsc{Gain}, \textsc{Loss}, \textsc{Stable}\},
\]
which summarizes coarse interaction changes between the robot gripper and its nearest object.

\subsection{Keyframe Selection}
We operate under a small keyframe budget $M$ and prioritize motion-salient frames. To detect salient events, we compute dense optical flow between consecutive frames and score each frame by its average flow magnitude. Keyframes are selected as local peaks in this score using temporal non-maximum suppression. If fewer than $M$ salient frames are identified, we supplement them with uniformly spaced frames to preserve contextual coverage.

We use \emph{dense} rather than sparse flow because our goal is not correspondence tracking, but a stable scene-wide saliency signal: manipulation failures often involve distributed motion of the arm, gripper, and object, while sparse keypoints can be unreliable in texture-poor or partially occluded robot scenes. The selector is modular and can be replaced by entropy-based or learned policies; our ablations compare motion-peak selection against uniform sampling.

\subsection{Per-Keyframe Perception}
\paragraph*{Open-Vocabulary Detection (OVD)}
We run an OVD module (e.g., GroundingDINO~\cite{groundingdino}) to detect objects of interest and robot arms/grippers. Detections are temporally linked across keyframes into short tracks (instance IDs), and timestamps $t_k$ are rendered as overlays on the RGB keyframes.

\paragraph*{Single-View Depth Estimation}
We estimate \emph{relative} depth per keyframe (e.g., Depth-Anything-V2~\cite{depth_anything_v2}) and associate depth statistics with each detection. We use depth only as a coarse ordering cue, not as metric geometry.

\paragraph*{Contact-Transition Proxy}
To provide a simple but informative interaction signal, we compute a coarse contact-transition token whenever a robot hand or gripper and a candidate object are both detected with high confidence in consecutive keyframes. Let $d_k$ denote the nearest-center distance between the gripper and its closest object at keyframe $k$, and let $\operatorname{IoU}_k$ denote the corresponding bounding-box IoU. With thresholds $\tau_{\text{IoU}},\tau_d>0$, we define
\[
\Delta \mathrm{IoU}_k = \mathrm{IoU}_{k+1}-\mathrm{IoU}_k,
\qquad
\Delta d_k = d_{k+1}-d_k.
\]
\[
\gamma_{k\rightarrow k+1} =
\begin{cases}
\textsc{Gain}, & \Delta \mathrm{IoU}_k \ge \tau_{\mathrm{IoU}},\ \Delta d_k \le -\tau_d,\\
\textsc{Loss}, & -\Delta \mathrm{IoU}_k \ge \tau_{\mathrm{IoU}},\ \Delta d_k \ge \tau_d,\\
\textsc{Stable}, & \text{otherwise.}
\end{cases}
\]
This token is intentionally coarse: it captures interaction trends that are often useful for failure analysis, without requiring force sensing or precise contact estimation.

\subsection{3D Scene Graph}
For each keyframe we build a local scene graph $G_k$ whose nodes are detections in $O_k$ with 3D centroids approximated from relative depth and camera geometry. We encode pairwise relations from the set
\[
\{\textsc{left\_of},\textsc{above},\textsc{in\_front\_of}\}
\]
using the sign and magnitude of centroid offsets with small tolerance thresholds. Local graphs $\{G_k\}_{k=1}^{M}$ are aggregated into a global graph by maintaining instance tracks across keyframes.

We use a \emph{coarse} 3D scene graph rather than a purely 2D graph because front/back ordering is often ambiguous in image coordinates alone, while relative depth provides enough signal to recover non-metric spatial ordering useful for failure diagnosis. Our goal is not full geometric reconstruction, but a compact relational scaffold that can disambiguate layout cues for the VLM.

\subsection{Robot Description}
We include a concise robot profile describing morphology (\#arms, \#grippers, end-effector types), sensors, workspace, and salient embodiment constraints. This allows the VLM to condition explanations and corrections on the robot platform and environment.

\subsection{Pseudo-BEV Schematic (Layout Prior)}
Photorealistic reconstructions are costly and not necessarily aligned with what current VLMs parse most reliably. We therefore render a schematic, non-metric top-down \emph{pseudo}-bird's-eye-view (pseudo-BEV) for each keyframe that externalizes relative layout while preserving identity consistency across modalities:
\begin{itemize}
\item fixed axes ($X$ right, $Z$ forward) with arrows;
\item one circle per tracked object, with radius proportional to confidence $s_j$;
\item the object class label and the same instance ID used in the RGB overlay;
\item overlaid timestamp $t_k$ and keyframe index.
\end{itemize}
Pseudo-BEVs are not metrically accurate maps; they are schematic layout cues intended to make spatial relationships easier for the VLM to read.

\subsection{KITE: Keyframe-Indexed Tokenized Evidence}
We serialize a compact context prefix that acts as a single front-end across all QA tasks. Let $\mathcal{T}$ denote the KITE context string:
{\small
\[
\begin{aligned}
\mathcal{T} =\;&
\underbrace{\texttt{[ROBOT]}~\text{short description}}_{\text{morphology, gripper, workspace}}
~\Vert~
\underbrace{\texttt{[PLAN]}~\text{high-level plan}}_{\text{optional task context}}
\\
&\Vert~
\underbrace{\texttt{[KF }i_k\texttt{ @ }t_k\texttt{]}}_{\text{timestamped keyframe tags}}
~\Vert~
\underbrace{\texttt{[CONTACT }k\!\rightarrow\!k{+}1\texttt{]}~\gamma_{k\rightarrow k+1}}_{\textsc{Gain}/\textsc{Loss}/\textsc{Stable}}
\\
&\Vert~
\underbrace{\texttt{[GLOBAL\_SCENE]}~\text{tracks \& relations}}_{\text{IDs consistent with RGB/pseudo-BEV}}.
\end{aligned}
\]
}
If plan steps are unavailable, the \texttt{[PLAN]} field is omitted.

\subsection{Prompting and Failure Localization}
For each question, we provide a compact image bundle consisting of RGB keyframe overlays and their corresponding pseudo-BEVs, and prepend $\mathcal{T}$ to the text prompt. We include a brief instruction stating that the pseudo-BEV is a \emph{schematic, not to scale}, and should be used only for relative layout reasoning.

For frame-level failure localization, we request \emph{strict JSON}:
\begin{quote}
\texttt{\{"candidates":[\{"frame\_num": INT, "confidence": FLOAT\}, ...]\}}
\end{quote}
with up to three candidates and confidence values in $[0,1]$. A simple parser extracts the top candidate, and subsequent analysis can then be aligned to that evidence frame.

\subsection{Narrative Summary}
Given $\mathcal{T}$ and a storyboard montage containing all selected keyframes and pseudo-BEVs, we prompt the VLM for a concise causal narrative that explicitly references keyframe IDs and timestamps, and proposes one high-level and one low-level correction. Since all perception is performed only on the selected keyframes, the overall cost scales linearly with $M$ and is independent of the original video length once $M$ is fixed.

\section{Experiments}
\label{sec:experiments}

We evaluate KITE on RoboFAC~\cite{robofac2025}, a large-scale benchmark for robotic failure analysis, using both quantitative and qualitative analyses. Our main question is whether the proposed front-end improves a strong off-the-shelf VLM \emph{without} task-specific training. Accordingly, our core comparison is vanilla Qwen2.5-VL versus KITE + Qwen2.5-VL, with the RoboFAC-tuned model and larger closed-source VLMs reported as reference baselines. We also provide ablations for pseudo-BEV and keyframe selection, and include qualitative rollouts from our lab robots---a RealMan dual-arm compound robot~\cite{realman2024compound} (DART) and ALOHA-2 Stationary~\cite{aloha2}---to illustrate transfer beyond the benchmark. RoboFAC contains only single-arm tasks, whereas our in-lab examples include dual-arm failures.

\subsection{Datasets and Tasks}
\paragraph*{RoboFAC}
RoboFAC~\cite{robofac2025} is a QA-style benchmark for robotic failure analysis containing both simulation and real-world sequences. It provides more than 60K training QA pairs from simulation, together with 10K simulated and 8K real-world QA pairs for testing. The benchmark defines eight question types: \emph{Task identification (TI)}, \emph{Task planning (TP)}, \emph{Failure detection (FD)}, \emph{Failure identification (FI)}, \emph{Failure locating (FL)}, \emph{Failure explanation (FE)}, \emph{High-level correction (HL)}, and \emph{Low-level correction (LL)}. In this paper, we report the seven tasks directly relevant to failure analysis and correction (TI, FD, FI, FL, FE, HL, LL); TP is used only as optional contextual information in KITE rather than as a primary evaluation target. We follow the official data splits and evaluation protocols where applicable.

\paragraph*{DART and ALOHA-2 (in-lab)}
We additionally test KITE qualitatively on in-lab sequences from DART and ALOHA-2. These examples are zero-shot with respect to our method and are intended to illustrate transfer to real dual-arm platforms and failure modes outside the single-arm benchmark setting.

\subsection{Backbones and Baselines}
\label{subsec:baselines}
We adopt Qwen2.5-VL~\cite{qwen2.5-vl} as the main backbone due to its strong vision-language capabilities and open-source availability. We additionally report results for Gemini-2.0~\cite{team2023gemini}, GPT-4o~\cite{openai2024gpt4o}, vanilla Qwen2.5-VL-3B and 7B models (without KITE; RGB keyframes only), the RoboFAC-7B model fine-tuned on RoboFAC~\cite{robofac2025}, our training-free KITE + Qwen2.5-VL, and KITE + Qwen2.5-VL further adapted with QLoRA.

\begin{figure*}[t]
\centering
\includegraphics[width=\linewidth]{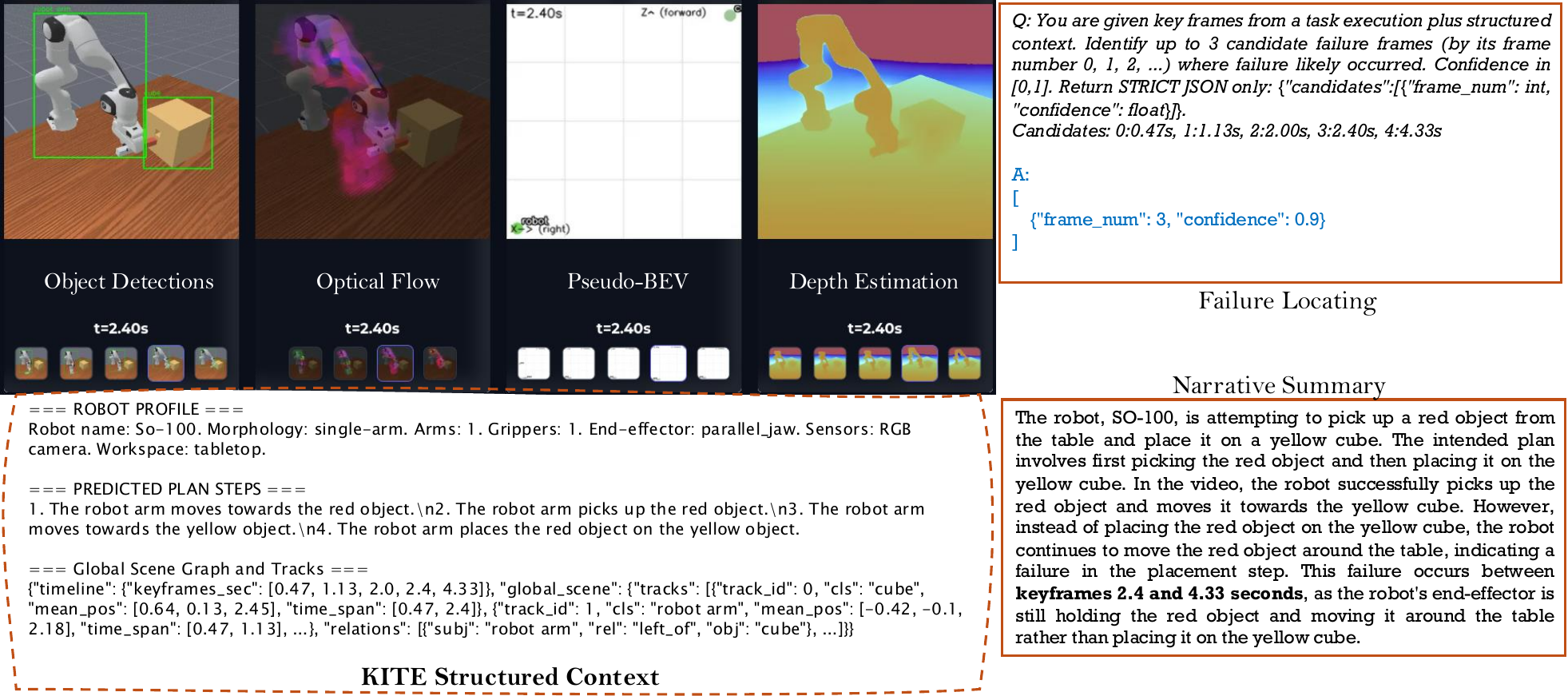}
\caption{\textbf{Qualitative results in simulation (RoboFAC dataset).} Each panel shows: RGB keyframe with object-detection overlays; optical-flow estimates; pseudo-BEV schematic (consistent object IDs; circle radius $\propto$ confidence; timestamp); and single-view depth estimates, all for the corresponding keyframes. We also illustrate a short structured-context excerpt, KITE's response to a failure-localization query, and a final narrative summary.}
\label{fig:qual_sim_robofac}
\end{figure*}

\subsection{Metrics}
For multiple-choice questions (FD, FI, FL), we report success rate. For free-language tasks (TI, FE, HL, LL), following~\cite{duan2024aha}, we report ROUGE-L F1 and Sentence-BERT cosine similarity between the generated answer and the reference answer.

\begin{table}[t]
\centering
\caption{Performance of multi-modal baseline models on the RoboFAC Benchmark~\cite{robofac2025}. Success rate for MCQ questions is reported (higher is better) for both simulation and real-world tasks.$\dag$ denotes the models that are finetuned on RoboFAC benchmark.}
\label{tab:robofac_mcq}
\setlength{\tabcolsep}{3pt}
\begin{tabular}{lcccccc}
\toprule
\multirow{2}{*}{Model} & \multicolumn{3}{c}{Simulation} & \multicolumn{3}{c}{Real-world} \\
\cmidrule(lr){2-4} \cmidrule(lr){5-7}
 & FD & FI & FL & FD & FI & FL \\
\midrule
Gemini-2.0 & 0.48 & \underline{0.27} & \textbf{0.75} & 0.60 & 0.11 & 0.18 \\
GPT-4o & \underline{0.64} & 0.21 & \underline{0.71} & \textbf{0.96} & \underline{0.43} & 0.52 \\
Qwen2.5-VL-3B & 0.38 & 0.04 & 0.51 & 0.04 & 0.03 & 0.07 \\
Qwen2.5-VL-7B & 0.52 & 0.26 & 0.22 & 0.83 & 0.38 & \underline{0.72} \\
\textbf{KITE + Qwen2.5-VL-7B} & \textbf{0.88} & \textbf{0.44} & 0.55 & \underline{0.84} & \textbf{0.43} & \textbf{0.74} \\
\midrule
\midrule
RoboFAC-7B$^\dag$ & 0.91 & 0.63 & \textbf{0.94} & 0.80 & 0.56 & 0.71 \\
\textbf{KITE+Qwen2.5-7B+QLoRA}$^\dag$ & \textbf{0.93} & \textbf{0.69} & 0.92 & \textbf{0.89} & \textbf{0.58} & \textbf{0.77} \\
\bottomrule
\end{tabular}
\end{table}

\subsection{Simulation and Real-world Results}
Table~\ref{tab:robofac_mcq} reports MCQ accuracy, and Table~\ref{tab:robofac_text} reports free-language results measured by ROUGE-L and Sentence-BERT similarity, for both simulation and real-world settings.

Training-free KITE substantially improves over vanilla Qwen2.5-VL-7B in simulation, with gains of $+36$ points on FD, $+18$ on FI, and $+33$ on FL. On real-world MCQ tasks, the gains over vanilla Qwen2.5-VL-7B are smaller but consistently positive ($+1$ on FD, $+5$ on FI, and $+2$ on FL). For free-language tasks, KITE improves ROUGE-L in all reported TI/FE/HL/LL settings and improves or closely matches Sentence-BERT similarity in nearly all cases. Applying QLoRA further improves performance across most reported dimensions and brings KITE close to, and in some cases beyond, the RoboFAC-tuned baseline.

\begin{table*}[t]
\centering
\caption{Performance of multi-modal baseline models on the RoboFAC Benchmark~\cite{robofac2025}. ROUGE-L and SBERT cosine similarity metrics (higher is better) are reported for free-language reasoning tasks for both simulation and real-world tasks. $\dag$ denotes the models that are finetuned on RoboFAC benchmark. }
\label{tab:robofac_text}
\setlength{\tabcolsep}{3pt}
\begin{tabular}{lcccccccc||cccccccc}
\toprule
\multirow{2}{*}{Model} & \multicolumn{4}{c}{Sim (ROUGE-L)} & \multicolumn{4}{c||}{Sim (SBERT Cosine)} & \multicolumn{4}{c}{Real (ROUGE-L)} & \multicolumn{4}{c}{Real (SBERT Cosine)} \\
\cmidrule(lr){2-5} \cmidrule(lr){6-9} \cmidrule(lr){10-13} \cmidrule(lr){14-17}
 & TI & FE & HL & LL & TI & FE & HL & LL & TI & FE & HL & LL & TI & FE & HL & LL \\
\midrule
Qwen2.5-VL-7B & 0.206 & 0.194 & 0.230 & 0.157 & 0.546 & 0.448 & 0.683 & 0.657 & 0.264 & 0.233 & 0.219 & 0.197 & 0.689 & 0.786 & 0.792 & 0.785 \\
\textbf{KITE + Qwen2.5-VL-7B} & 0.295 & 0.248 & 0.241 & 0.190 & 0.680 & 0.829 & 0.798 & 0.779 & 0.300 & 0.252 & 0.223 & 0.232 & 0.696 & 0.832 & 0.791 & 0.804 \\
\midrule
\midrule
RoboFAC-7B$^\dag$ & 0.323 & 0.299 & 0.301 & 0.245 & 0.701 & 0.842 & 0.808 & 0.794 & 0.337 & 0.361 & 0.228 & 0.305 & 0.722 & 0.856 & 0.798 & 0.813 \\
\textbf{KITE+Qwen2.5-7B+QLoRA}$^\dag$ & 0.326 & 0.314 & 0.302 & 0.296 & 0.698 & 0.845 & 0.806 & 0.803 & 0.338 & 0.365 & 0.229 & 0.313 & 0.724 & 0.860 & 0.798 & 0.815 \\
\bottomrule
\end{tabular}
\end{table*}

\subsection{Ablations}
\label{subsec:ablations}
We isolate the contribution of pseudo-BEV and the keyframe selector on real-world tasks, as shown in Table~\ref{tab:ablation_sim}. In the \emph{$\downarrow$ pseudo-BEV} setting, we remove only the pseudo-BEV images while keeping the RGB keyframes and all serialized text tokens unchanged. In the \emph{uniform keyframe} setting, we replace motion-based selection with uniformly spaced keyframes while keeping the rest of KITE fixed.

Removing pseudo-BEV reduces performance most clearly on failure explanation, where FE drops by 0.05 ROUGE-L, and also hurts FD/FI/FL. Replacing motion-based keyframes with uniform sampling causes a larger degradation overall, especially for questions that depend on identifying when and where the failure first becomes visible.

\begin{table}[t]
\centering
\caption{Ablation study of our method. The $\downarrow$ indicates the feature is removed. Success rate for MCQ and ROUGE-L metric for other question dimensions are reported for real-world tasks.}
\label{tab:ablation_sim}
\setlength{\tabcolsep}{4pt}
\begin{tabular}{lccccccc}
\toprule
\multirow{2}{*}{Config} & \multicolumn{3}{c}{success rate} & \multicolumn{4}{c}{ROUGE-L} \\
\cmidrule(lr){2-4} \cmidrule(lr){5-8}
 & FD & FI & FL & TI & FE & HL & LL  \\
\midrule
Full (KITE) & 0.84 & 0.43 & 0.74 & 0.300 & 0.252 & 0.223 & 0.232 \\
$\downarrow$ pseudo-BEV & 0.81 & 0.37 & 0.70 & 0.302 & 0.202 & 0.221 & 0.228 \\
uniform keyframe & 0.69 & 0.33 & 0.56 & 0.298 & 0.189 & 0.217 & 0.190 \\
\bottomrule
\end{tabular}
\end{table}

\subsection{Qualitative Analyses}
For a sequence from the \texttt{PegInsertionSide} simulation task in RoboFAC, Fig.~\ref{fig:qual_sim_robofac} shows the selected keyframes, object detections, optical-flow estimates, pseudo-BEV renderings, depth estimates, a short excerpt of the KITE context, the failure-localization output, and the final narrative summary. The example illustrates how KITE makes the evidence chain legible to the VLM.

\begin{figure*}[t]
\centering
\includegraphics[width=\textwidth]{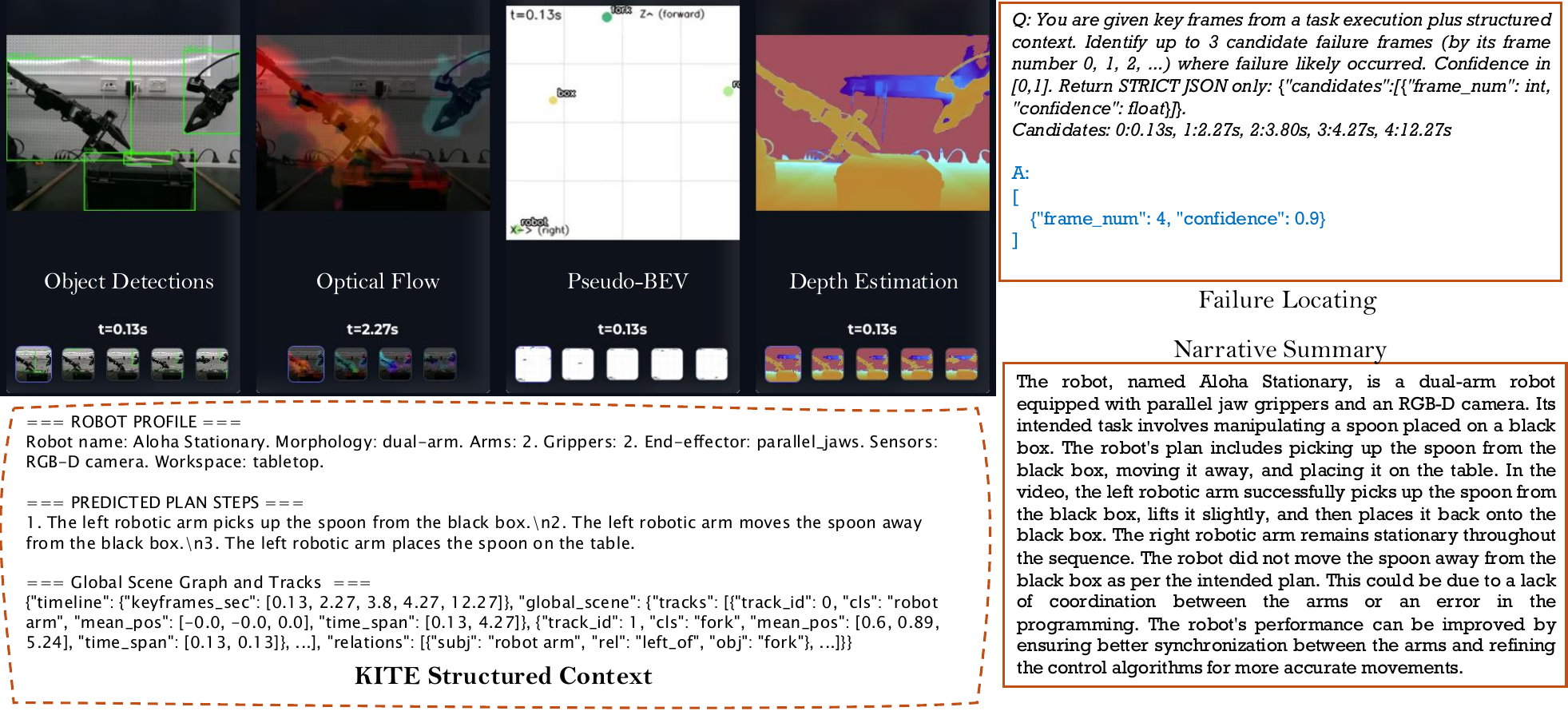}
\caption{\textbf{Qualitative results in real-world (ALOHA-2).} Each panel shows: RGB keyframe with object-detection overlays; optical-flow estimates; pseudo-BEV schematic (consistent object IDs; circle radius $\propto$ confidence; timestamp); and single-view depth estimates, all for the corresponding keyframes. We also illustrate a short structured-context excerpt, KITE's response to a failure-localization query, and a final narrative summary.}
\label{fig:qual_real_aloha}
\end{figure*}

We also present the same intermediate representations and outputs for real-world sequences recorded in our lab: DART in Fig.~\ref{fig:teaser}, and ALOHA-2 in Fig.~\ref{fig:qual_real_aloha}. In Fig.~\ref{fig:qual_real_aloha}, the object is dropped during a dual-arm handover. The generated explanation explicitly ties the failure to the robot embodiment and the observed sequence, illustrating the value of including robot-profile information in the structured context.

\subsection{Implementation Details}
\label{sec:impl}
\paragraph{Keyframes.}
We use up to $M=8$ keyframes per video. This budget was chosen to balance temporal coverage against the multimodal context length that can be passed to the VLM. Keyframes are selected primarily from motion-salient peaks detected via optical flow; if fewer than $M$ salient frames are found, we add uniformly spaced frames to preserve coverage. All images are resized to $512\times512$ for VLM input.

\paragraph{Optical Flow, OVD, Depth, and Contact.}
We compute per-frame mean optical-flow magnitude using dense flow~\cite{farneback2003two} to obtain a scene-wide motion saliency score for keyframe proposal. We use GroundingDINO~\cite{groundingdino} (Swin-T backbone) for open-vocabulary detection, capped at five detections per keyframe. For monocular depth, we use Depth-Anything-V2-Large~\cite{depth_anything_v2}; to reduce extreme outliers, we suppress depth values beyond the 0.8 quantile bound. Contact-transition tokens use IoU and nearest-center trends across adjacent keyframes, as described in Section~\ref{sec:method}.

\paragraph{Contact-Transition Thresholds.}
For the contact-transition proxy, we set $\tau_{\text{IoU}}=0.1$ for bounding-box overlap and $\tau_d=0.15$ for nearest-center distance.

\paragraph{Pseudo-BEV.}
Pseudo-BEV schematics are rendered on a $256\times256$ white canvas with $X/Z$ axes, projected semantic dots for tracked objects, confidence-scaled circle radii clipped to $[r_{\min}=3, r_{\max}=10]$ pixels, class labels, and OCR-friendly timestamps.

\paragraph{VLM Calls.}
For each QA, we provide $2\times M$ images: the RGB keyframes and their corresponding pseudo-BEVs. The text prompt consists of the KITE prefix, a short instruction explaining that the pseudo-BEV is a schematic top-down layout used for \emph{relative} spatial reasoning, and the target question.

\paragraph{QLoRA.}
To study how well the proposed evidence representation transfers under lightweight adaptation, we also fine-tune the VLM with QLoRA~\cite{qlora}. We use rank 8, 4-bit quantization, one epoch, and a learning rate of $1\times10^{-5}$, with the LLM backbone and merger parameters unfrozen. All training and evaluation are run on a single NVIDIA A6000 GPU.


%
\section{Limitations and Future Work}
KITE deliberately favors compact, interpretable evidence over full geometric fidelity. It relies on open-vocabulary detection and monocular relative depth, which can struggle with small, occluded, reflective, or visually ambiguous objects. Its contact-transition proxy captures coarse interaction trends rather than precise force events. Likewise, the current scene graph is intentionally lightweight: it uses a small set of coarse relations and omits potentially useful predicates such as \textsc{on\_top\_of} and \textsc{inside}, so the system is better suited to diagnosing high-level spatial inconsistencies than precise geometric deviations. The pseudo-BEV is non-metric and flattens vertical structure, which further limits low-level geometric analysis.

The keyframe selector is also a simplification. Motion saliency can miss low-motion or very brief failures, and identity tracking across sparse keyframes can switch in cluttered scenes. More broadly, our quantitative evaluation is centered on RoboFAC, while the DART and ALOHA-2 studies are qualitative; broader cross-benchmark evaluation and user-facing assessment of explanation quality remain important future directions. Finally, results still depend on the reasoning quality of the chosen VLM backend. Future work can therefore explore stronger perception modules, richer relation vocabularies, multi-view layout cues, adaptive keyframe policies, and broader human-centered evaluation.

\section{Conclusion}
We introduced KITE, a training-free, keyframe-indexed, pseudo-BEV-grounded front-end that converts long robot-execution videos into compact, interpretable tokenized evidence for VLMs. By combining object-overlaid keyframe RGBs, schematic layout cues, robot-profile information, contact-transition tokens, and serialized scene relations, KITE provides a structured interface for failure detection, identification, localization, explanation, and correction. On RoboFAC, KITE substantially improves a strong vanilla VLM baseline in the training-free setting and remains competitive with task-tuned alternatives, while a lightweight QLoRA adaptation yields further gains. Qualitative results on DART and ALOHA-2 suggest that the representation transfers beyond a single benchmark and extends naturally to dual-arm real-world failures.

%









\bibliographystyle{IEEEtran}
\bibliography{references}

\end{document}